\documentclass[letterpaper, 10 pt, conference]{ieeeconf}  %

\IEEEoverridecommandlockouts                              %

\usepackage{graphics} %
\usepackage{epsfig} %
\usepackage{mathptmx} %
\usepackage{times} %
\usepackage{amsmath} %
\usepackage{amssymb}  %

\usepackage[hidelinks]{hyperref}
\usepackage[capitalize]{cleveref}
\usepackage{booktabs}
\usepackage{multirow}
\usepackage[table]{xcolor}
\usepackage{array}
\usepackage{xspace}
\usepackage{eso-pic}
\IfFileExists{dsfont.sty}{\usepackage{dsfont}}{\let\mathds\mathbf}
\usepackage{pifont}
\usepackage{microtype}
\usepackage{caption}
\newcommand{\parsection}[1]{\par\noindent\textbf{#1}}
\definecolor{darkgreen}{rgb}{0.1, 0.5, 0.1}
\newcommand{\cmark}{\color{darkgreen}\ding{51}}%
\newcommand{\xmark}{\color{red}\ding{55}}%

\newcommand{\eg}{\textit{e.g.}\xspace}

\title{\LARGE \bf
3D Open-Vocabulary Detection via\\
Visual Alignment and Generative Captions
}

\author{
Thomas E. Huang, Alex Zihao Zhu, Hao Xiang, Tianyi Cai, Jingwei Ji,\\
Yuliang Zou, Jiafan Yu, Xin Huang, and Mingxing Tan\\Waymo%
}

\begin{document}

\maketitle
\AddToShipoutPictureBG*{%
  \AtPageLowerLeft{%
    \hspace*{\dimexpr 1in + \oddsidemargin\relax}%
    \raisebox{0.20in}{%
      \parbox[b]{\textwidth}{%
        \scriptsize
        \noindent\copyright~2026 IEEE. Personal use of this material is permitted. Permission from IEEE must be obtained for all other uses, in any current or future media, including reprinting/republishing this material for advertising or promotional purposes, creating new collective works, for resale or redistribution to servers or lists, or reuse of any copyrighted component of this work in other works.%
      }%
    }%
  }%
}
\thispagestyle{empty}
\pagestyle{empty}

\begin{abstract}

Open-vocabulary 3D object detection remains a significant challenge for outdoor scenes, with existing methods often struggling to effectively transfer the rich semantic knowledge from powerful 2D Vision-Language Models (VLMs) into a 3D context.
We propose a novel, two-part framework for this task.
First, we introduce a direct, object-level distillation method.
We generate class-agnostic 3D proposals to source regions from the entire scene, without relying on any manual labels.
We leverage a spatially-aware VLM to extract a semantic embedding from each proposal's projected 2D bounding box and supervise our 3D model to predict this embedding on a per-instance basis.
Second, to enhance zero-shot capability and inject domain-specific information, we employ a powerful Multimodal Large Language Model (MLLM) as an image captioner to generate rich textual captions for each of these proposals.
These captions are integrated into the training objective to directly align the 3D features with these rich semantics, creating a composite loss that combines direct feature distillation with caption-based semantic enrichment.
Our experiments validate the effectiveness of our approach, establishing a new state of the art in 3D open-vocabulary detection on the Waymo Open Dataset and nuScenes.
\end{abstract}

\section{Introduction}
\label{sec:intro}

Accurate 3D object detection is foundational for safety-critical applications like autonomous driving.
Current state-of-the-art 3D detectors, which rely on LiDAR point clouds~\cite{wang2023dsvt, fan2023fully} and multi-view images~\cite{li2024bevformer, wang2023unitr}, have achieved remarkable performance on challenging benchmarks~\cite{Sun_2020_CVPR, caesar2020nuscenes}.
However, these methods are fundamentally limited: they can only identify and localize objects that belong to a pre-defined set of categories seen during training.
In real-world environments, systems constantly encounter novel objects and long-tail concepts, rendering closed-set detectors unreliable.
When a closed-set detector fails to identify a novel obstacle such as debris or an unconventional vehicle, it poses a direct threat to downstream motion planning and collision avoidance systems.

The emerging field of open-vocabulary 3D object detection addresses this limitation by aiming to detect arbitrary objects described by natural language.
To achieve this open-set capability, prevailing approaches typically rely on knowledge distillation from large-scale pre-trained 2D Vision-Language Models (VLMs)~\cite{radford2021learning}.
These methods transfer the VLM's rich semantic features to the 3D detector, either by projecting 2D features to 3D points/voxels for alignment~\cite{peng2023openscene}, or by using 2D open-vocabulary detections~\cite{zhou2022detecting, liu2024grounding} as pseudo-labels for 3D localization~\cite{zhang2024opensight, etchegaray2024find}.
This enables the 3D model to predict an embedding vector that can be compared against arbitrary text queries (\eg, object names) at inference time.

\begin{figure}[t]
    \centering
    \includegraphics[width=0.9\linewidth]{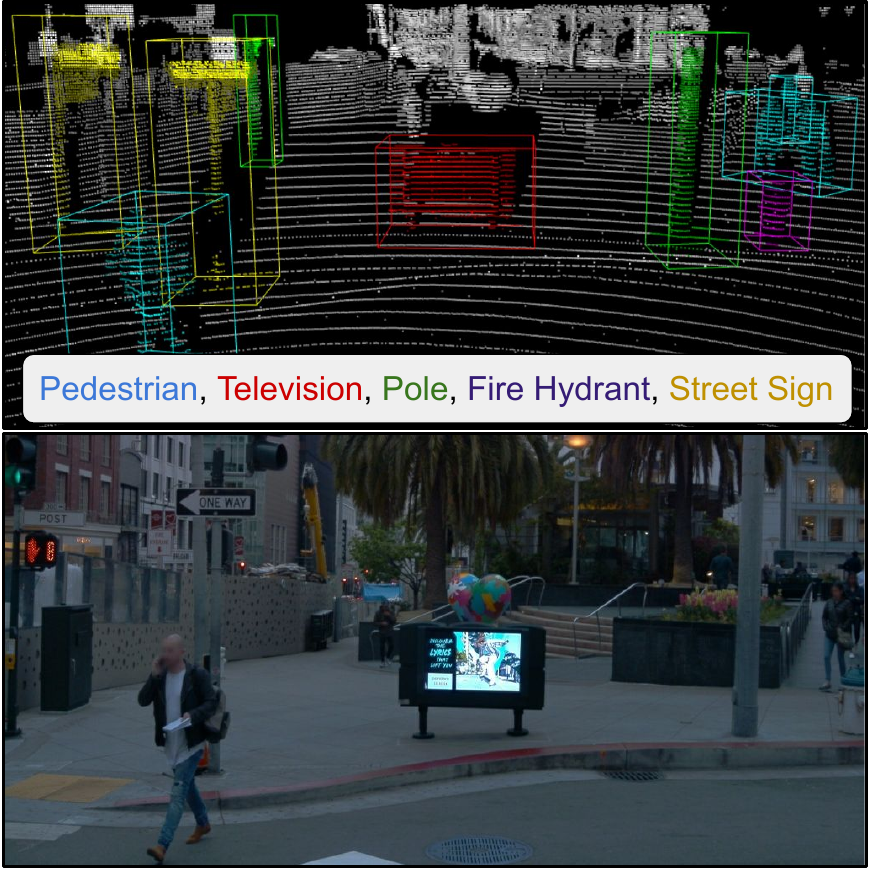}
    \vspace{-0.06in}
    \caption{\textbf{3D Open-Vocabulary Detection Results.} Given text prompts for each class, our detector can retrieve corresponding objects in 3D, without using any ground-truth data during training. Our detector can even generalize to novel object classes like TVs.}
    \vspace{-0.24in}
    \label{fig:teaser}
\end{figure}

The primary challenge in this domain is bridging the knowledge gap between the feature-rich 2D domain, where vast Vision-Language Models (VLMs) like CLIP~\cite{radford2021learning} have acquired broad open-world knowledge, and the sparse, 3D domain.
Directly aligning sparse, unorganized, 3D features with dense, pre-trained 2D image features is notoriously noisy.
Additionally, the 2D VLMs are pre-trained on web-scale data, which are general, but may create significant domain gaps when applied to specific domains.
3D autonomous driving environments are often cluttered, occluded, and have highly varied-perspective views.
Relying solely on this visual-only alignment is insufficient; the 2D visual features are often semantically ambiguous and lack the granularity to distinguish between fine-grained categories (\eg, `trailer' vs. `truck').
This forces the 3D model to learn a similarly coarse representation, limiting its generalization.

In this work, we introduce a novel, end-to-end framework that effectively transfers and grounds open-world semantic knowledge to a 3D detector.
First, our Visual Alignment stage (\cref{method:vis-align}) creates a robust visual foundation.
To overcome the noise and sparsity of 3D-to-2D projection, we distill knowledge in an object-centric manner.
We first generate class-agnostic 3D box proposals using DBSCAN~\cite{dbscan} on the raw point cloud, allowing us to source distillation targets from all regions of the scene, including background and potential novel objects, without relying on any ground-truth labels or prior knowledge.
We aggregate 2D VLM features at the object level and use this stable visual signal to supervise the underlying 3D voxel features.
This efficiently distills the VLM's rich visual knowledge into the 3D backbone, creating a feature space that is visually discriminative.

While this VLM distillation provides general open-world knowledge, the original image embeddings are pre-trained to cover a broad range of captions, and may miss domain specific context. To this end, we propose a second Language Alignment stage (\cref{method:lang-align}) which complements this visual-alignment and instills fine-grained semantic understanding.
We leverage a powerful Multimodal Large Language Model (MLLM) as an image captioner and domain adapter.
Using the same DBSCAN-generated 3D regions, we generate high-fidelity, descriptive captions directly from the 3D dataset's image patches. These captions are then embedded into the VLM embedding space with the VLM text encoder.
By aligning the visually-grounded 3D features (from Stage 1) to these domain-specific text embeddings, we overcome the semantic ambiguity of 2D visual features and explicitly force it to learn the nuanced semantics of the 3D environment (\eg, ``a white sedan stopped in traffic"). 
In short, our loss combines the breadth of pre-trained image embedding distillation with the depth of per-object caption embedding distillation.

Our contributions are summarized as follows:
\begin{enumerate}
    \item We introduce a Visual Alignment strategy that uses class-agnostic proposals to distill object-level 2D VLM features into their corresponding 3D voxels, enabling robust feature learning from the entire scene without manual labels.
    \item We propose a Language Alignment strategy that uses an MLLM to generate high-fidelity, contextual text captions for these same scene regions, grounding the 3D features in rich, descriptive language.
    \item We achieve state-of-the-art results for open-vocabulary 3D object detection on the challenging Waymo Open Dataset and nuScenes benchmarks.
\end{enumerate}

\begin{figure*}[t]
    \vspace*{0.04in}
    \centering
    \includegraphics[width=0.95\linewidth]{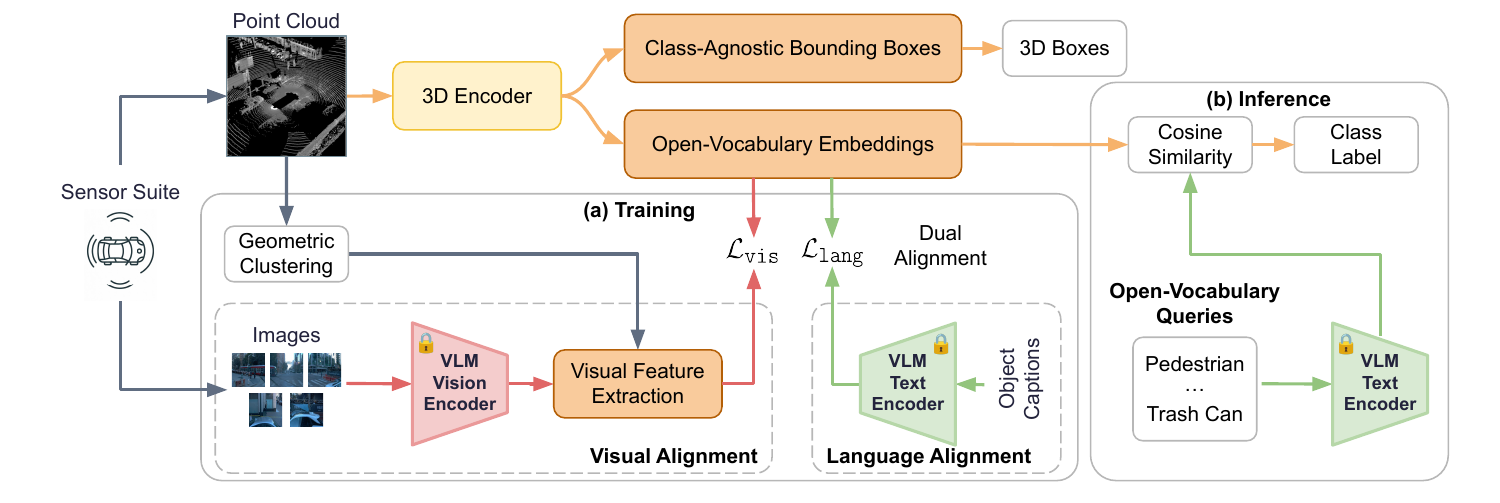}
    \vspace{-0.12in}
    \caption{\textbf{Overall Pipeline.} During (a) training, our detector is supervised with three losses: class-agnostic bounding box loss $\mathcal{L}_{\texttt{det}}$ along with two distillation losses, Visual Alignment loss $\mathcal{L}_{\texttt{vis}}$ and Language Alignment loss $\mathcal{L}_{\texttt{lang}}$, for feature alignment. During (b) inference, we directly use the trained detector to predict boxes and open-vocabulary embeddings. Text-based queries are encoded by the VLM text encoder, and cosine similarity between the embeddings and queries are used to assign the final class label.}
    \label{fig:pipeline}
    \vspace{-0.16in}
\end{figure*}

\section{Related Work}

\subsection{3D Closed-Vocabulary Object Detection}
Mainstream 3D object detectors can be broadly categorized by their data representation.
Early works~\cite{qi2017pointnet++, shi2019pointrcnn} operated directly on raw point clouds, but were often computationally expensive.
The dominant paradigm shifted to voxel-based methods~\cite{zhou2018voxelnet, yan2018second}, which discretize the scene into a regular grid, enabling efficient 3D sparse convolutions.
This was further refined by hybrid point-voxel architectures~\cite{shi2020pv} and efficient anchor-free heads like CenterPoint~\cite{yin2021center}.
More recently, Transformer-based detectors~\cite{bai2022transfusion, sun2022swformer, fan2023fully, wang2023dsvt, wang2023unitr, li2024bevformer} have achieved state-of-the-art performance.
Despite their high accuracy, these methods are fundamentally "closed-set": they are trained with a fixed vocabulary (\eg, car, pedestrian) and cannot detect any object outside this pre-defined list.

\subsection{Vision-Language Modeling}
The dominant paradigm in training Vision-Language Models (VLMs) has been contrastive learning on web-scale image-text pairs, pioneered by CLIP~\cite{radford2021learning} and ALIGN~\cite{jia2021scaling}.
These models learn a joint embedding space, enabling powerful zero-shot image-level classification.
However, this image-level supervision limits their spatial granularity, making them suboptimal for dense prediction tasks like detection.
Subsequent works have focused on adapting these models for region-level understanding, either by fine-tuning on detection data~\cite{li2022grounded, zhong2022regionclip}, creating dense feature maps~\cite{zhou2022denseclip}, or incorporating self-supervised objectives for spatial awareness, such as TIPS~\cite{maninis2025tips}.
A separate and parallel class of models, Multimodal Large Language Models (MLLMs), has emerged by connecting vision encoders to Large Language Models (LLMs)~\cite{liu2023visual, dai2023instructblip, comanici2025gemini}.
These models excel at complex reasoning and generative tasks like captioning.

\subsection{Open-Vocabulary Object Detection}
Open-vocabulary detection was first popularized in 2D.
Early methods~\cite{gu2021open} focused on distilling knowledge from a VLM like CLIP~\cite{radford2021learning} into a region-based detector~\cite{ren2015faster} by aligning region-of-interest (RoI) features with the VLM's text embeddings.
More recent approaches have developed end-to-end transformer frameworks~\cite{minderer2022simple} or improved performance by pre-training on large-scale grounded image-text data~\cite{li2022grounded, liu2024grounding}.
Extending this to 3D introduces the significant challenge of bridging the 2D-3D modality gap.
One line of work applies 2D open-vocabulary models to multi-view images and ``lifts" the resulting 2D boxes into 3D space~\cite{zhang2024opensight, etchegaray2024find, najibi2023unsupervised, chow2025ov}.
However, the open vocabulary needs to be known before training, and extension to new categories require data regeneration and retraining.
A more direct approach aligns 3D features directly with the VLM's embedding space.
This is often done by projecting entire 3D point clouds into 2D views~\cite{zhang2022pointclip, zhu2023pointclip, lin2023l3det} or by distilling 2D VLM features into 3D point or voxel features~\cite{peng2023openscene, chen2023clip2scene}.
Our work follows this latter distillation-based approach.

\section{Method}

In this section, we introduce our method for 3D open-vocabulary object detection.
We first provide the setting, and then we describe key components of our method, Visual Alignment and Language Alignment.
The full training and inference pipeline is shown in \cref{fig:pipeline}.

\subsection{Setting}
\label{method:background}
For LiDAR-based object detection in outdoor scenes, the input is a 3D point cloud $\mathcal{P}$, represented as a set of $N$ points, $\mathcal{P} = \{p_i\}_{i=1}^N$, where each point $p_i \in \mathbb{R}^{3+F}$ includes its 3D coordinates $(x, y, z)$ and optionally $F$ additional features (\eg, intensity, color).
The detector processes this input point cloud and is tasked with predicting $M$ objects, $G = \{(b_j, c_j)\}_{j=1}^M$, where $b_j$ is the 3D bounding box and $c_j$ is the class label.
A 3D bounding box is parameterized by its center $(x, y, z)$, dimensions $(l, w, h)$, and yaw orientation $\theta$.
Thus, $b_j = (x_j, y_j, z_j, l_j, w_j, h_j, \theta_j)$.
In standard detection, $c_j$ is from a fixed set of object classes, $C_{\text{train}} = \{c_1, \dots, c_K\}$.
Along with the LiDAR point cloud, a set of $L$ synchronized multi-view images covering different perspectives $\mathcal{I}=\{I_l\}_{l=1}^L$ with corresponding extrinsic $R_l$ and intrinsic $K_l$ camera parameters are provided.
We only use the images during training.

In 3D open-vocabulary object detection, the detector is tasked with predicting objects from an ``open" vocabulary, including categories unseen during training.
Standard classification does not work in this setting.
Typically, language is used as an intermediary~\cite{minderer2022simple, liu2024grounding}, and the detector predicts an open-vocabulary embedding for each object, rather than the standard classification objective.
This enables classification through comparison against arbitrary text queries during inference, thereby generalizing to unseen categories.

\begin{figure*}[t]
    \vspace*{0.06in}
    \centering
    \includegraphics[width=0.95\linewidth]{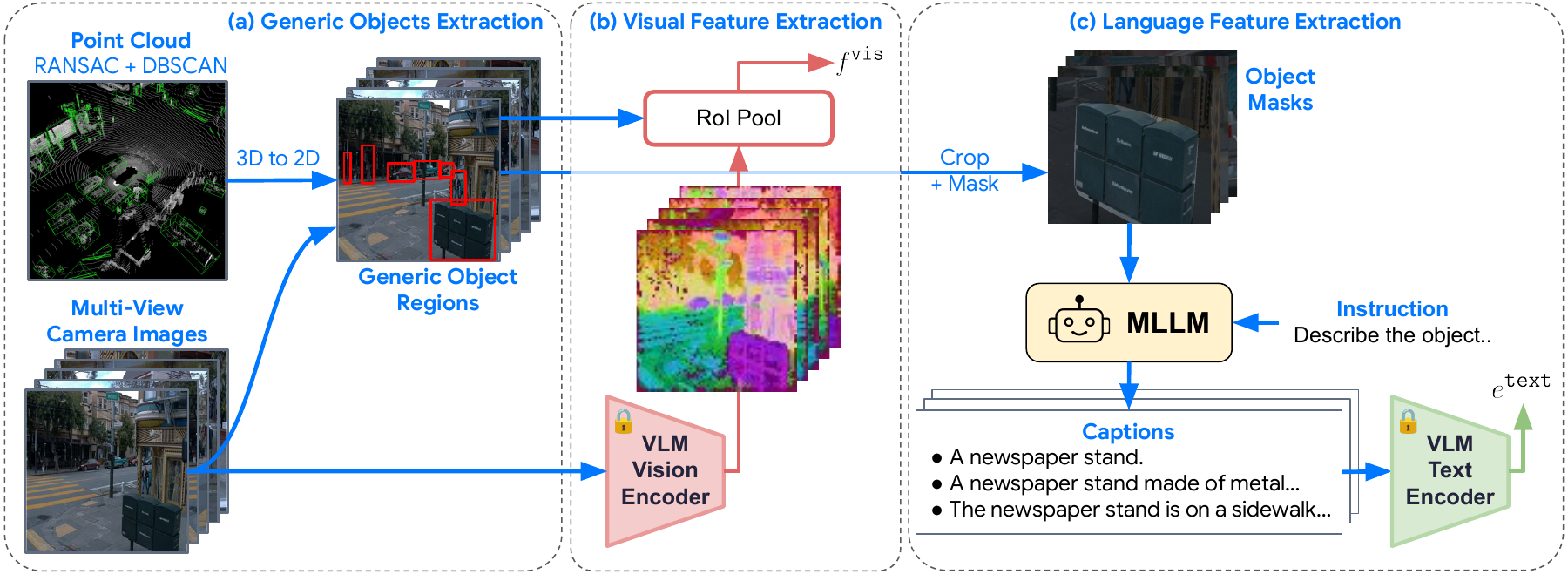}
    \vspace{-0.06in}
    \caption{\textbf{Data Generation Pipeline.} (a) Generic Objects Extraction: We use RANSAC and DBSCAN on the raw point cloud to generate class-agnostic 3D proposals across the entire scene, which are then projected to all 2D camera views. (b) Visual Feature Extraction: The proposals are used to RoI pool features from the VLM vision encoder, creating the visual distillation targets $f^{\texttt{vis}}$. (c) Language Feature Extraction: The proposals are used to crop and mask the images and then sent to a MLLM for captioning, which are then encoded by the VLM text encoder to create the language distillation targets $e^{\texttt{text}}$.}
    \label{fig:data-pipeline}
    \vspace{-0.16in}
\end{figure*}

\begin{figure}[t]
    \centering
    \includegraphics[width=0.8\linewidth]{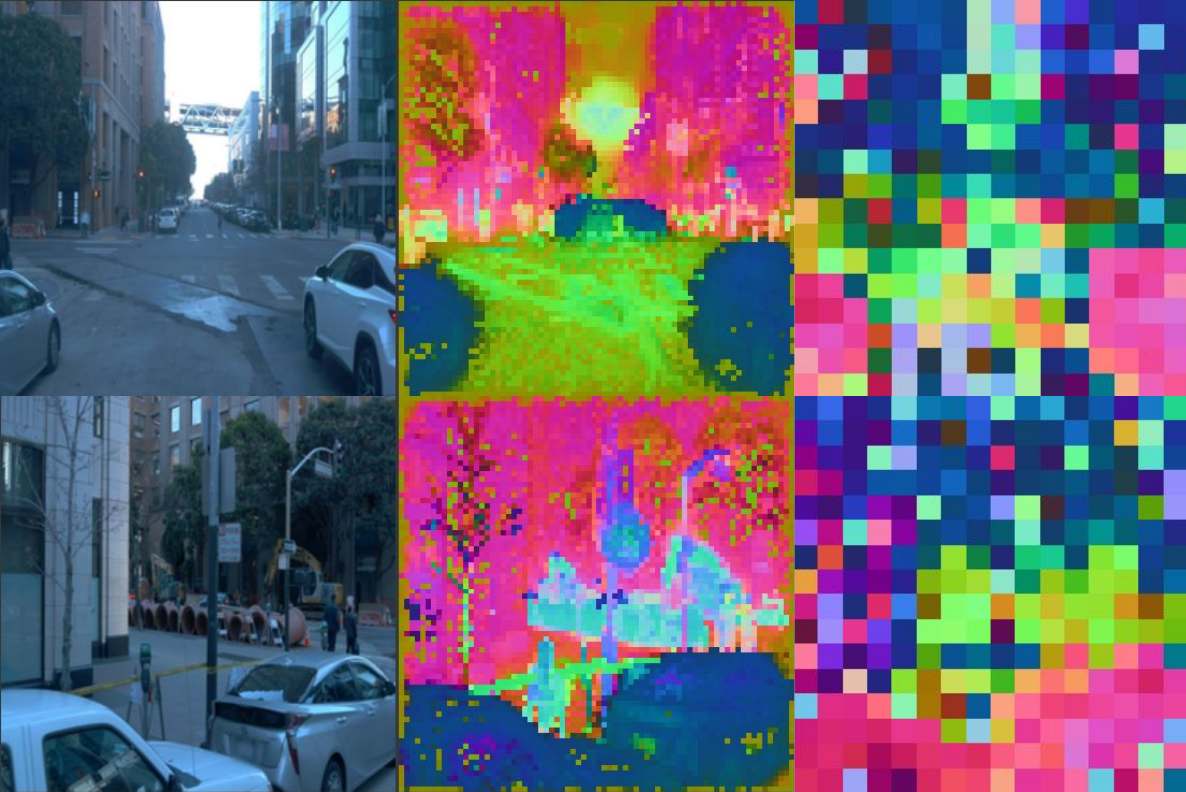}
    \vspace{-0.06in}
    \caption{\textbf{TIPS Features.} Comparison of visual features from TIPS~\cite{maninis2025tips} (middle) and CLIP~\cite{radford2021learning} (right). TIPS produce features that are higher resolution and more spatially aligned.}
    \label{fig:tips-feats}
    \vspace{-0.16in}
\end{figure}

\subsection{Visual Alignment}
\label{method:vis-align}

The first component of our method focuses on establishing a visual-semantic link between the 3D features and the rich, pre-trained knowledge embedded in a VLM.
We aim to directly distill such knowledge into the 3D features to enable open-vocabulary classification directly in 3D.

\parsection{3D Feature Extraction.}
Our alignment framework is designed to be agnostic to the 3D backbone.
For our experiments, we build upon a Transformer-based point cloud detector, SWFormer~\cite{sun2022swformer}.
Given an input point cloud, SWFormer first applies voxelization and then uses a series of sparse window Transformer blocks for feature processing and multi-scale feature fusion to produce voxel features $F^{\texttt{3D}}\in\mathbb{R}^{N_v\times D}$.

\parsection{Generic Objects Extraction.}
A primary challenge in aligning 3D features with 2D VLM features is establishing a robust correspondence.
The most direct approach is to perform point-to-pixel alignment~\cite{peng2023openscene}.
This is achieved by projecting each 3D point $p_i$ onto the 2D image plane using the given extrinsic $R_l$ and intrinsic $K_l$ camera parameters:
\begin{equation}
\label{eq:proj}
    p^{\texttt{2D}}_{li} = K_l \cdot R_l \cdot p_i.
\end{equation}
However, this point-to-pixel strategy is quite brittle, especially for outdoor scenes.
A 3D point can be easily assigned an incorrect 2D semantic feature due to minor calibration errors or small occlusions (\eg, a pole in front of a vehicle), which in turn introduces noise into the training signal.

To mitigate the noise inherent in point-to-pixel alignment, we adopt an object-centric distillation strategy without relying on manual labels.
We deviate, however, from the standard practice of lifting 2D detection boxes~\cite{zhang2024opensight, etchegaray2024find},
which necessitate a fixed list of categories during training for prompting the 2D detectors and limits true open-vocabulary generalization.
Instead, we treat object discovery as a geometric problem rather than a semantic one.
We employ RANSAC~\cite{ransac} to separate the ground plane and DBSCAN~\cite{dbscan} to cluster the remaining points into distinct entities.
This strategy generates a comprehensive set of class-agnostic 3D proposals covering the entire scene.
For each 3D proposal $b'_{j}$, we compute its 2D projection $b^{\texttt{2D}}_{lj}$ across all camera views using Equation~\ref{eq:proj}.
This process is visualized in \cref{fig:data-pipeline}~(a).

Projecting all 3D geometric proposals to 2D indiscriminately can still generate low-quality pairs.
This introduces noise, for example, by forcing the model to align a 3D object with a 2D crop of its occluder.
To ensure high-fidelity 2D-3D correspondence, we apply a rigorous filtering process.
First, we perform view frustum and depth filtering.
We discard any 2D projected boxes that are fully outside the image bounds, are too small or large, or whose corresponding 3D box is closer than 5m or further than 50m.
Second, we perform occlusion filtering to resolve ambiguity in the camera perspective.
We calculate the Intersection over Area (IoA) for overlapping boxes and remove any box that has a high IoA with another box and is further away from the camera, as it is presumed to be occluded.

\parsection{Vision-Language Model.}
Our object-centric distillation requires a VLM that can produce spatially-aware, high-resolution feature maps.
Standard VLMs like CLIP~\cite{radford2021learning} are not ideal for this, as they are trained for image-level tasks with a global image-text alignment objective.
Consequently, prior works~\cite{cao2023coda} using CLIP must rely on an inefficient process: running separate forward passes on individual image crops to extract visual features for each object.
This approach, however, is context-poor, as it discards all surrounding scene information.
To overcome this, we use TIPS~\cite{maninis2025tips}.
TIPS is explicitly designed for spatial awareness, leveraging a DINO-like self-supervised loss~\cite{caron2021emerging} and a CLIP-style loss to produce high-resolution, spatially-aligned visual features (\cref{fig:tips-feats}).
This key property allows us to extract meaningful 2D features from a dense feature map of the entire image.
Furthermore, each patch retains global scene context due to its attention mechanism.
For our pipeline, we encode each image at full resolution to obtain high-resolution feature maps $\mathcal{E}_V(I_l)=F_l^{\texttt{TIPS}}$ for $l=1,\dots,L$, where $\mathcal{E}_V$ is the TIPS vision encoder and $F^{\texttt{TIPS}}\in\mathbb{R}^{L\times H\times W \times C}$.
These features can be computed offline and do not incur any cost during training.

\parsection{Visual Distillation.}
For each 2D box, we apply RoI pooling on the VLM features to obtain a single, robust 2D visual feature vector,
\begin{equation}
    f^{\texttt{vis}}_j=\frac{1}{|L^v_j|}\sum_{l\in L^v_j}\left(\texttt{RoIPool}(b^{\texttt{2D}}_{lj}, F_l^{\texttt{TIPS}})\right),
\end{equation}
where $L^v_j$ represent the set of camera indices with valid 3D-to-2D projections for box $b'_j$.
The resulting features represent the visual VLM features that act as the targets for distillation.

For visual feature alignment, we first use a lightweight open-vocabulary embedding head to predict an embedding for each voxel, $f^{\texttt{emb}}=\texttt{MLP}(F^{\texttt{3D}})$.
We then take the predictions corresponding to voxels within each object's 3D bounding box and
minimize the cosine distance between the predicted and the target embeddings:
\begin{equation}
    \mathcal{L}_{\texttt{vis}} = \frac{1}{M'} \sum_{j=1}^{M'} \frac{1}{N_{\texttt{v}}} \sum_{i=1}^{N_{\texttt{v}}} \mathds{1}(v_i \in b'_j) \left(1 - \frac{f_i^{\texttt{emb}} \cdot f^{\texttt{vis}}_j}{\|f_i^{\texttt{emb}}\| \|f^{\texttt{vis}}_j\|}\right),
\end{equation}
where $M'$ is the number of object boxes after filtering, $N_{\texttt{v}}$ is the total number of voxels, and $\mathds{1}$ is an indicator function representing the indices of all voxels that belong to object $j$.
This loss forces the 3D features to encode rich visual information consistent with VLM's 2D representation.

\subsection{Language Alignment}
\label{method:lang-align}

While visual alignment transfers visual knowledge from pre-trained VLMs, the VLM image encoder is trained on a large corpus of image-text pairs, and so the image-text similarity may not be sufficiently discriminative for any given text prompt.
Additionally, the image embeddings suffer from visual ambiguity (\eg, distinguishing a truck from a semi-truck solely by its 2D patch) and the limited textual diversity of standard class labels.
To overcome this and instill a richer semantic understanding, our second component focuses on bridging this vision-language gap through leveraging contextually-rich language generated by a Multimodal Large Language Model (MLLM) image captioner.
We posit that aligning to MLLM-generated captions, which are rich in contextual and descriptive attributes (\eg, `a white sedan stopped in traffic'), forces the 3D model to learn a more nuanced semantic space than aligning to the fixed, iconic visual concepts learned by the VLM (\eg, a general `car').

\begin{figure}[t]
    \vspace*{0.06in}
    \centering
    \includegraphics[width=0.99\linewidth]{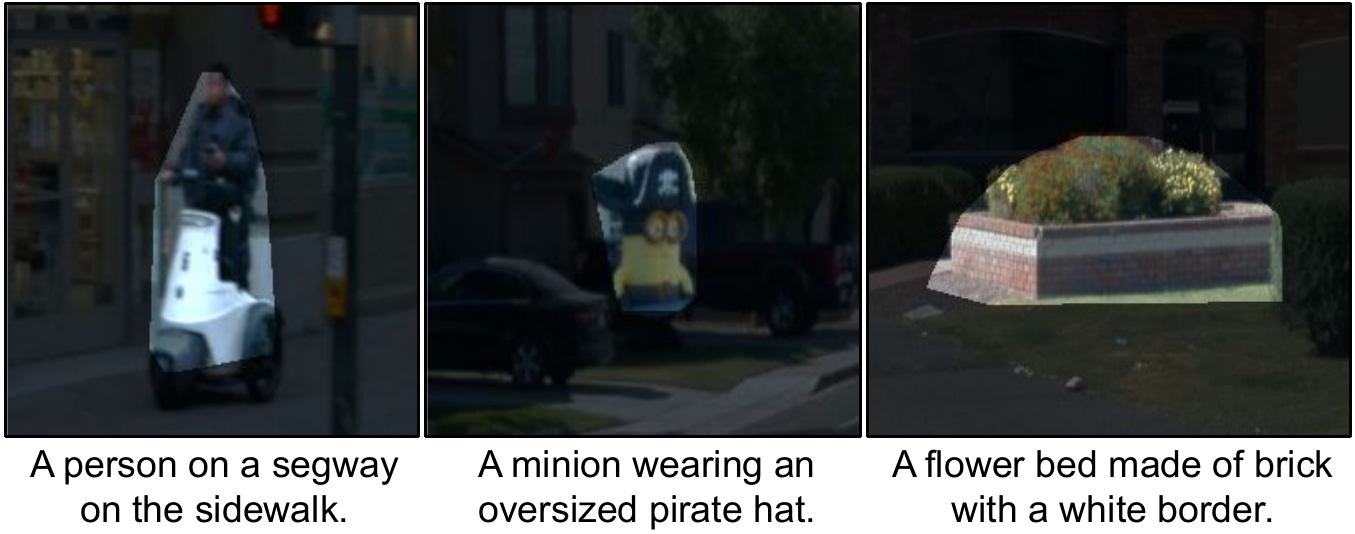}
    \vspace{-0.2in}
    \caption{\textbf{Example Captions.} We obtain detailed descriptions of rare objects in the Waymo Open Dataset.}
    \label{fig:gemini-captions}
    \vspace{-0.16in}
\end{figure}

\parsection{Captioning.}
We use the publicly available Gemini 2.5 Flash~\cite{comanici2025gemini}
 as our captioning MLLM.
Using the same set of class-agnostic 3D proposals $b'_{j}$ from \cref{method:vis-align}, we project points within each 3D bounding box to the 2D image planes and compute convex hulls of the LiDAR points inside the box, representing visible regions of each object.
After filtering and occlusion handling, we convert each convex hull to a binary mask and crop the camera images around each mask.
To incorporate more context information into each crop, we enlarge each crop by a constant factor and dim the brightness of the background region.
We pair each crop with an instruction, asking Gemini to generate a set of text descriptions, each corresponding to a different attribute: object type, appearance, surroundings, and action. During training, we randomly sample one of these attributes to be used as the caption for each object.
The full captioning pipeline is shown in \cref{fig:data-pipeline}~(c).
Examples of image inputs and partial responses from Gemini (appearance description) are shown in \cref{fig:gemini-captions}.

This process generates high-fidelity, contextual text descriptions $T_j$ based on the MLLM's immense world knowledge, which provide richer semantic signals than using coarse category names, critical for grounding fine-grained and novel concepts.
Additionally, this removes the need for a pre-defined category list for novel objects, which many previous methods rely on~\cite{zhang2024opensight, etchegaray2024find, chow2025ov}.
We encode these descriptions using the VLM's pre-trained text encoder $\mathcal{E}_T$ to obtain object text embeddings $\mathcal{E}_T(T_j)=e_j^{\texttt{text}}$.

\parsection{Text Distillation.}
To align the open-vocabulary embeddings from the detector with the MLLM-generated text embeddings, we first use a lightweight adapter to project the visual-aligned features to the text feature space $e^{\texttt{emb}}=\texttt{MLP}\left( f^{\texttt{emb}} \right)$.
Then, we minimize the cosine distance:

\begin{equation}
    \mathcal{L}_{\texttt{lang}} = \frac{1}{M'} \sum_{j=1}^{M'} \frac{1}{N_{\texttt{v}}} \sum_{i=1}^{N_{\texttt{v}}} \mathds{1}(v_i \in b'_j) \left(1 - \frac{e_i^{\texttt{emb}} \cdot e^{\texttt{text}}_j}{\|e_i^{\texttt{emb}}\| \|e^{\texttt{text}}_j\|}\right).
\end{equation}
This loss explicitly grounds the 3D object features in the semantic space defined by human language, improving the model's ability to match the 3D observations to open-vocabulary textual queries during inference.

\begin{table*}[t]
\vspace*{0.06in}
\footnotesize
\centering
\caption{
Open-vocabulary 3D object detection on WOD (ViLGOD~\cite{fruhwirth2024vision} setting).
No ground-truth labels are used during training.
AP is computed with IoU=0.4 at difficulty level L2.
$\dagger$ indicates evaluation using ground-truth class labels, results from \cite{fruhwirth2024vision}.
}
\vspace{-0.06in}
\scalebox{0.99}{
\begin{tabular}{@{\extracolsep{\fill}} l | c c c c | c c c c}
\toprule
\multirow{2}{*}{Method} & \multicolumn{4}{c|}{BEV AP} & \multicolumn{4}{c}{3D AP} \\
\cmidrule(lr){2-5} \cmidrule(lr){6-9} & Vehicle & Pedestrian & Cyclist & mAP & Vehicle & Pedestrian & Cyclist & mAP \\
\midrule
\multicolumn{9}{c}{\textit{Standard range (100m x 40m)}} \\
DBSCAN~\cite{dbscan}~$\dagger$ & 18.4 & 0.2 & 0.1 & 6.2 & 4.8 & 0.0 & 0.0 & 1.6 \\
LSF~\cite{Deng_2023_WACV}~$\dagger$ & 10.9 & 0.0 & 0.2 & 3.7 & 7.4 & 0.0 & 0.0 & 2.5 \\
OYSTER~\cite{zhang2023towards}~$\dagger$ & 56.2 & 0.0 & 0.0 & 18.7 & 20.4 & 0.0 & 0.0 & 6.8 \\
LISO~\cite{Baur2024ECCV}~$\dagger$ & \textbf{69.5} & 5.5 & 2.2 & 25.7 & 54.3 & 3.7 & 1.6 & 19.9 \\
ViLGOD~\cite{fruhwirth2024vision} & 49.0 & 16.8 & 7.6 & 24.5 & 44.8 & 14.1 & 7.4 & 22.1 \\
ViLGOD-CP~\cite{fruhwirth2024vision} & 64.4 & \textbf{38.8} & 7.5 & 36.9 & \textbf{62.4} & \textbf{35.9} & 7.5 & 35.3 \\
\midrule
Class-Agnostic Baseline & 8.9 & 8.0 & 1.0 & 6.0 & 7.8 & 6.3 & 0.8 & 5.0 \\
\textbf{Ours} & 63.7 & 36.0 & \textbf{42.3} & \textbf{47.3} (\textcolor{darkgreen}{+10.3}) & 59.3 & 32.5 & \textbf{42.3} & \textbf{44.7} (\textcolor{darkgreen}{+9.4}) \\
\midrule
\multicolumn{9}{c}{\textit{Extended range (160m x 160m)}} \\
ViLGOD~\cite{fruhwirth2024vision} & 27.2 & 12.3 & 4.7 & 14.7 & - & - & - & - \\
ViLGOD-CP~\cite{fruhwirth2024vision} & 29.5 & 23.9 & 4.6 & 19.3 & - & - & - & - \\
\midrule
Class-Agnostic Baseline & 7.1 & 2.6 & 0.9 & 3.5 & 5.7 & 2.1 & 0.8 & 2.9 \\
\textbf{Ours} & \textbf{38.1} & \textbf{27.4} & \textbf{27.7} & \textbf{31.1} (\textcolor{darkgreen}{+11.8}) & \textbf{36.4} & \textbf{26.7} & \textbf{27.4} & \textbf{30.2} \\
\bottomrule
\end{tabular}
}
\label{tab:wod-vilgod}
\vspace{-0.12in}
\end{table*}

\begin{table}[t]
\centering
\small
\caption{
Open-vocabulary 3D object detection on WOD (standard range) (UP-VL~\cite{najibi2023unsupervised} setting).
No ground-truth labels are used during training.
VRU combines pedestrians and cyclists into a single class.
AP is computed with IoU=0.4 at difficulty level L2.
* indicates class-agnostic AP is used.
}
\vspace{-0.06in}
\scalebox{0.9}{
\begin{tabular}{@{\extracolsep{\fill}} l c | c c c}
\toprule
\multirow{2}{*}{Method} & \multirow{2}{*}{VLM} & \multicolumn{3}{c}{3D AP} \\
\cmidrule(lr){3-5} & & Vehicle & VRU & mAP \\
\midrule
Clustering~\cite{najibi2022motion} & - & - & - & 11.6* \\
Clustering~\cite{najibi2022motion} & OpenSeg~\cite{ghiasi2022scaling} & 16.1 & 10.0 & 13.1 \\
Auto labels~\cite{najibi2023unsupervised} & OpenSeg~\cite{ghiasi2022scaling} & 30.2 & 14.7 & 22.4 \\
UP-VL~\cite{najibi2023unsupervised} & OpenSeg~\cite{ghiasi2022scaling} & 52.0 & 19.7 & 35.8 \\
\midrule
Class-Agnostic & - & 7.8 & 3.2 & 5.5 \\
\textbf{Ours} & TIPS~\cite{maninis2025tips} & \textbf{59.3} & \textbf{29.7} & \textbf{44.5} (\textcolor{darkgreen}{+8.7}) \\
\bottomrule
\end{tabular}
}
\label{tab:wod-upvl}
\vspace{-0.12in}
\end{table}

\subsection{Model Details}
We provide additional details regarding our model.

\parsection{Localization.}
Following \cite{sun2022swformer}, we utilize a CenterNet detection head for regressing 3D bounding boxes from voxel features.
Since we do not have access to a pre-defined vocabulary during training,
we use a single detection head for all objects, and we modify the classification branch to output a binary foreground/background prediction rather than classify into object classes.
As 3D localization training requires precise bounding boxes,
this component is trained with a separate set of standard, amodal 3D boxes for supervision,
rather than the DBSCAN boxes used for alignment.

\parsection{Training.}
We train our detector in an end-to-end manner by jointly optimizing all objectives:
\begin{equation}
    \mathcal{L}_{\texttt{total}} = \mathcal{L}_{\texttt{det}} + \lambda_{\texttt{vis}}\mathcal{L}_{\texttt{vis}} + \lambda_{\texttt{lang}}\mathcal{L}_{\texttt{lang}},
\end{equation}
where $\mathcal{L}_{\texttt{det}}$ is the standard loss for the CenterNet detection head, including heatmap, size, and offset regression losses,
and $\lambda_{\texttt{vis}}$ and $\lambda_{\texttt{lang}}$ are used to weight the alignment losses.

Our method does not require images during inference time, and the proposed alignment components are only used during training.
Thus, the only additional computation are the lightweight open-vocabulary embedding heads, which are a tiny fraction of the overall detector.

\parsection{Inference.}
To classify objects into open-vocabulary categories, we rely on the open-vocabulary embeddings from the model.
During inference, we take textual descriptions of categories of interest $T_c$ and encode them using the VLM text encoder $\mathcal{E}_T(T_c)=e_c$.
These descriptions can be any free-form text (\cref{fig:qual-res}).
Then, we simply compute the cosine similarity between the predicted and the provided text embeddings and assign the label with the highest score.

\section{Experiments}

\subsection{Experimental Setup}

\begin{table*}[t]
\vspace*{0.06in}
\centering
\small
\caption{Open-vocabulary 3D object detection on nuScenes. For fair comparison, all methods use \textbf{ground-truth boxes} for training (no class labels). We report the overall mAP, NDS, and individual class APs. Results with $\dagger$ are taken from \cite{zhang2024opensight}.
}
\vspace{-0.06in}
\scalebox{0.84}{
\begin{tabular}{l c | c c | c c c c c c c c c c}
\toprule
Method & VLM & mAP & NDS & Car & Truck & C.V. & Bus & Trailer & Barrier & Bicycle & T.C. & Motor. & Ped. \\
\midrule
OV-PointCLIP~\cite{zhang2022pointclip}~$\dagger$ & CLIP~\cite{radford2021learning} & 0.1 & 5.0 & 0.0 & 0.0 & 0.0 & 0.0 & 0.0 & 0.0 & 0.7 & 0.2 & 0.0 & 0.1 \\
OV-PointCLIPv2~\cite{zhu2023pointclip}~$\dagger$ & CLIP~\cite{radford2021learning} & 2.8 & 10.9 & 0.0 & 2.7 & 0.0 & 0.0 & 0.0 & 22.0 & 0.0 & 0.2 & 0.0 & 3.2 \\
OpenSight~\cite{zhang2024opensight}~$\dagger$ & G.DINO~\cite{liu2024grounding} & 34.2 & 39.0 & 53.2 & 20.1 & \textbf{7.4} & 20.5 & \textbf{0.7} & 50.1 & \textbf{37.8} & 52.6 & \textbf{39.1} & 60.8 \\
\midrule
Class-Agnostic Baseline & - & 5.4 & 23.1 & 25.5 & 0.0 & 0.0 & 0.0 & 0.0 & 3.2 & 0.0 & 0.2 & 0.0 & 25.5 \\
\textbf{Ours} & TIPS~\cite{maninis2025tips} & \textbf{36.9} (\textcolor{darkgreen}{+2.7}) & \textbf{40.3} (\textcolor{darkgreen}{+1.3}) & \textbf{71.2} & \textbf{21.3} & 0.0 & \textbf{31.9} & 0.4 & \textbf{63.1} & 19.5 & \textbf{54.2} & 37.6 & \textbf{70.0} \\
\bottomrule
\end{tabular}
}
\label{tab:nuscenes_results}
\vspace{-0.06in}
\end{table*}

\begin{table}[t]
\centering
\small
\caption{Ablation study on our proposed alignment losses. Each loss alone significantly improves classification performance, and the joint loss provides significantly improved results.}
\vspace{-0.06in}
\scalebox{0.98}{
\begin{tabular}{c c c c c c}
\toprule
\multirow{2}{*}{Visual} & \multirow{2}{*}{Language} & \multicolumn{4}{c}{BEV AP} \\
\cmidrule(lr){3-6}
 & & Vehicle & Pedestrian & Cyclist & mAP \\
\midrule
\xmark & \xmark & 8.9 & 8.0 & 1.0 & 6.0 \\
\cmark & \xmark & \textbf{63.8} & 32.7 & 37.9 & 44.8 \\
\xmark & \cmark & 63.0 & 34.2 & 39.9 & 45.7 \\
\midrule
\cmark & \cmark & 63.7 & \textbf{36.0} & \textbf{42.3} & \textbf{47.3} \\
\bottomrule
\end{tabular}
}
\label{tab:abl-align}
\vspace{-0.08in}
\end{table}

\parsection{Dataset.}
We conduct experiments on the Waymo Open Dataset (WOD)~\cite{Sun_2020_CVPR} and nuScenes~\cite{caesar2020nuscenes}.
WOD includes 1000 sequences, split into 798 for training and 202 for validation.
The dataset uses five cameras, covering a 270-degree view around the vehicle.
WOD contains annotations for three object categories: vehicle, pedestrian, and cyclist.
The nuScenes dataset contains 1000 driving sequences, split into 700 for training, 150 for validation, and 150 for testing.
Each sequence is 20 seconds in length and annotated at 2Hz.
It also has images taken from six cameras, covering 360-degrees around the vehicle.
nuScenes provides annotations for ten object categories for the detection task.

\parsection{Evaluation Settings.}
The core contribution of our work is the open-vocabulary alignment pipeline, which is agnostic to the underlying 3D localization method.
To isolate the performance of our alignment strategy,
we conduct experiments using boxes that are available to us, including unsupervised, amodal pseudo-labels from~\cite{najibi2023unsupervised} on WOD and ground-truth (GT) boxes (no class labels) on nuScenes.
These boxes are used to supervise our class-agnostic detection branch.

For WOD, we follow existing open-vocabulary protocols~\cite{fruhwirth2024vision, najibi2023unsupervised} and evaluate zero-shot 3D detection performance without utilizing any GT labels.
We report 3D and BEV Average Precision (AP) at an IoU threshold of 0.4 and difficulty level 2 (all objects).
We use two range settings: the standard range (100m × 40m around the ego-vehicle) and the extended range (160m x 160m).
As \cite{najibi2023unsupervised} and ~\cite{fruhwirth2024vision} use slightly different evaluation protocols, we compare with each separately.
We compare with methods that support open-vocabulary inference (ViLGOD~\cite{fruhwirth2024vision}, UP-VL~\cite{najibi2023unsupervised}) and unsupervised detection methods with GT label assignment (OYSTER~\cite{zhang2023towards}, LISO~\cite{Baur2024ECCV}).

For nuScenes, we found that most existing open-vocabulary results were skewed by the performance of the box generation pipeline. In order to focus on open-vocabulary classification, we compare with OpenSight~\cite{zhang2024opensight} and its baselines that report performance using GT boxes (but not classes) for training.
We report nuScenes mAP, NDS, and individual class APs.

\subsection{Main Results}
In this section, we present the main results of our method.
For the class-agnostic baseline, we use the same class-agnostic detector as our method to generate boxes, but assign class labels of all classes to each box.

\parsection{Waymo Open Dataset.}
The results on the WOD dataset are shown in \cref{tab:wod-vilgod} and \cref{tab:wod-upvl}.
Compared with ViLGOD~\cite{fruhwirth2024vision}, our method obtains competitive performance on vehicles and pedestrians but significantly better results on cyclists, resulting in an overall 3D mAP improvement of 9.4 points.
Previous methods perform worse for cyclists due to rarity and similarity to pedestrians, while our detector can successfully detect them with high accuracy.
In addition, we demonstrate our method's generalization to long ranges by evaluating on the extended range setting from ViLGOD~\cite{fruhwirth2024vision}.
Our method achieves an mAP of 31.1, significantly outperforming the ViLGOD by 11.8 points.
Notably, we see massive gains in detecting vulnerable road users (VRUs), with Cyclist AP improving from 4.6 to 27.7 (+23.1).

Compared to UP-VL~\cite{najibi2023unsupervised}, our method achieves large performance gains across both vehicles and VRUs, improving overall mAP by 8.7 points.
As we use the same auto labels, this demonstrate that our distillation strategy is more effective in transferring world knowledge to the 3D detector compared to the point-to-pixel strategy employed in UP-VL.

\parsection{nuScenes.}
The results on the nuScenes dataset are presented in \cref{tab:nuscenes_results}.
OpenSight~\cite{zhang2024opensight} relies on Grounding DINO~\cite{liu2024grounding} for open-vocabulary classification by providing the 2D detector with prompts of class names both during training for feature alignment and inference for feature integration.
This results in limited generalization to truly novel classes where the names are not known beforehand.
Our method does not have such a limitation.
Despite this, our method achieves 36.9 mAP, a 2.7 points improvement over OpenSight, demonstrating the effectiveness of our alignment protocol.

\begin{figure*}[t]
    \centering
    \includegraphics[width=0.99\linewidth]{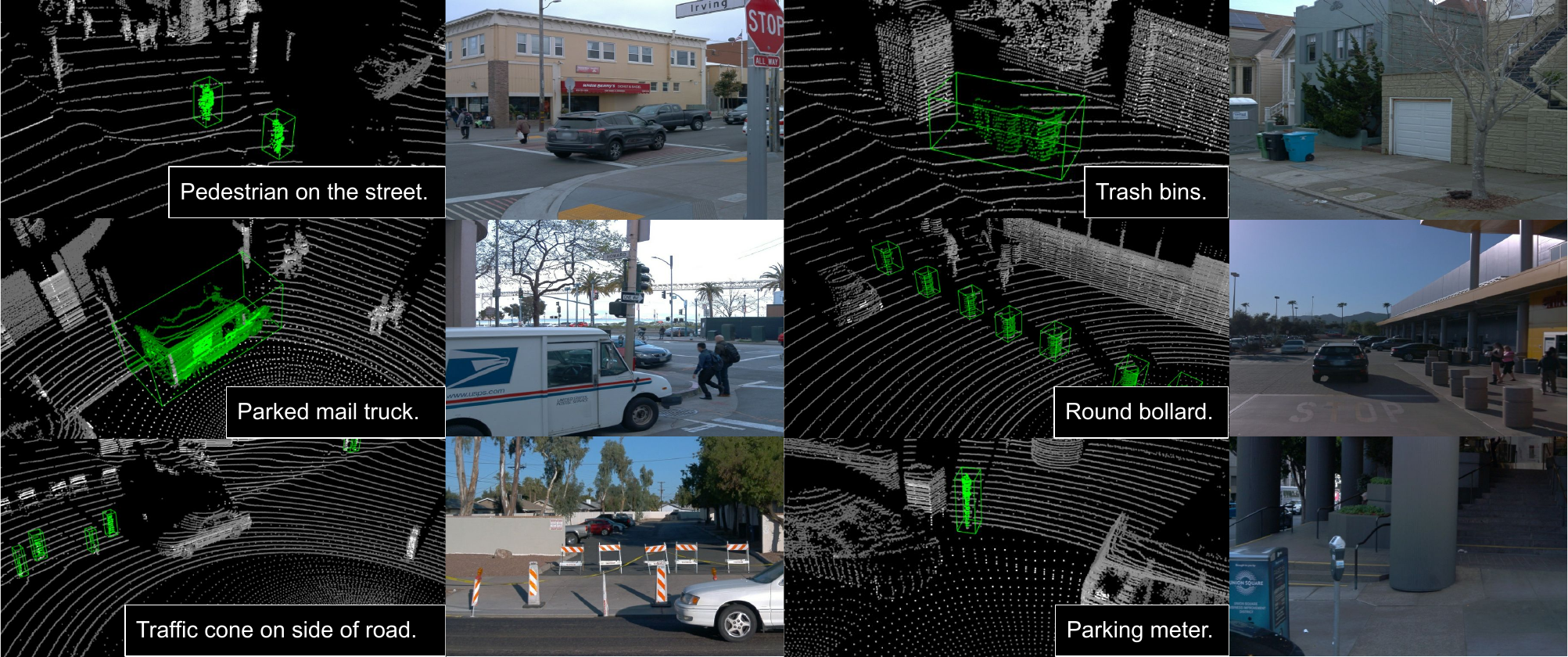}
    \vspace{-0.06in}
    \caption{\textbf{Qualitative Zero-Shot Results.} Our model successfully detects objects when given diverse, fine-grained textual descriptions. }
    \label{fig:qual-res}
    \vspace{-0.14in}
\end{figure*}

\subsection{Ablation Studies}
We conduct comprehensive ablation studies on the WOD validation set to verify the contribution of each component.
We use the same setting as \cref{tab:wod-vilgod} with the standard range setting for all ablation experiments.

\parsection{Component Analysis.}
We first analyze the impact of our two proposed alignment stages in \cref{tab:abl-align}.
Without any alignment, the detector is unable to classify objects, achieving a low mAP of 6.0.
Using only Visual Alignment yields a strong score of 44.8 mAP, confirming that distilling spatially-aware TIPS features is effective.
Using only Language Alignment also performs well at 45.7 mAP, showing that generative captions alone can provide strong supervision.
Combining both streams achieves the best performance (47.3 mAP).
This indicates that the two modalities are complementary: visual alignment provides fine-grained spatial cues, while language alignment provides richer semantic grounding.

\begin{table}[t]
\centering
\small
\caption{Ablation study on the Visual Alignment strategy. Ground-Truth Box denotes using GT instead of DBSCAN boxes for Visual Alignment.}
\vspace{-0.06in}
\scalebox{0.90}{
\begin{tabular}{l c c c c}
\toprule
\multirow{2}{*}{Strategy} & \multicolumn{4}{c}{BEV AP} \\
\cmidrule(lr){2-5}
 & Vehicle & Pedestrian & Cyclist & mAP \\
\midrule
Baseline & 8.9 & 8.0 & 1.0 & 6.0 \\
\midrule
+ Full 2D Box & 53.1 & 25.6 & 28.4 & 35.7 \\
+ Visible Points Filtering & 56.0 & 27.9 & 33.2 & 39.0 \\
+ Occlusion Handling & \textbf{63.8} & \textbf{32.7} & \textbf{37.9} & \textbf{44.8} \\
+ Ground-Truth Box & 51.0 & 23.0 & 22.8 & 32.3 \\
\bottomrule
\end{tabular}
}
\label{tab:abl-vis-align}
\vspace{-0.18in}
\end{table}

\parsection{Visual Alignment Strategy.}
We now ablate the different design choices of our Visual Alignment component in \cref{tab:abl-vis-align}.
Simply projecting all 3D bounding boxes to 2D and performing distillation with all correspondences can enable open-vocabulary detection but only achieves 35.7 mAP.
Projecting points within each 3D bounding box instead to filter boxes without visible points can improve the performance to 39.0 mAP.
Finally, using occlusion handling to filter out boxes that are overlapping with others can further improve detection accuracy to 44.8 mAP.
We also show that performing distillation in all regions is beneficial for learning by comparing with using ground-truth regions only for Visual Alignment.
The results indicate that limiting to such regions hurts performance significantly, resulting in only 32.3 mAP.

\parsection{Language Alignment Strategy.}
We also ablate the different design choices of our Language Alignment component in \cref{tab:abl-lang-align}.
Specifically, we investigate the impact of using different attributes as captions for distillation.
Using only textual descriptions about each object's type, or each object's appearance can already achieve strong performance of 45.6 and 45.5 mAP, respectively.
Using object types can help better classify cyclists, while using object appearance descriptions further enhances accuracy on more common vehicles and pedestrians.
Using all available captions (type, appearance, surroundings, action) gives us the benefits of multiple aspects, providing a balance that improves overall mAP.
We further compare with utilizing ground-truth class names as the text target for Language Alignment by distilling only in ground-truth regions.
The results show that the lack of diversity in captions hurt the overall performance.

\begin{table}[t]
\centering
\small
\caption{Ablation study on the captions used for Language Alignment. Ground-Truth Classes denotes using GT class names instead of generated captions for Language Alignment.}
\vspace{-0.06in}
\scalebox{0.9}{
\begin{tabular}{l c c c c}
\toprule
\multirow{2}{*}{Strategy} & \multicolumn{4}{c}{BEV AP} \\
\cmidrule(lr){2-5}
 & Vehicle & Pedestrian & Cyclist & mAP \\
\midrule
Baseline & 8.9 & 8.0 & 1.0 & 6.0 \\
\midrule
Object Type & 62.6 & 33.6 & 40.6 & 45.6 \\
Object Appearance & \textbf{63.5} & \textbf{35.0} & 38.4 & 45.5 \\
All Captions & 63.0 & 34.2 & \textbf{39.9} & \textbf{45.7} \\
Ground-Truth Classes & 49.7 & 26.6 & 27.0 & 34.4 \\
\bottomrule
\end{tabular}
}
\label{tab:abl-lang-align}
\vspace{-0.16in}
\end{table}

\parsection{Visualization.}
We provide qualitative results in \cref{fig:qual-res} to demonstrate our detector's ability to detect objects with arbitrary text queries.
Without any ground-truth annotations during training, our detector is able to handle diverse and fine-grained textual descriptions during inference.

\addtolength{\textheight}{-0.3cm}   %

\section{Conclusion}

In this work, we presented a novel 3D open-vocabulary framework that learns robust, open-world representations by aligning 3D features to both spatially-aware VLMs and generative MLLM captions.
We use class-agnostic geometric clustering to source proposals directly from the point cloud and a dual-stream multi-modal distillation framework to align these proposals with the VLM features.
Our experiments demonstrate state-of-the-art open-vocabulary detection performance on the Waymo Open Dataset and nuScenes,
validating that our approach can better align 3D geometry with 2D large-scale pre-trained models.

{
    \small
    \bibliographystyle{IEEEtran}
    \bibliography{main}
}

\end{document}